\documentclass[10pt,twocolumn,letterpaper]{article}

\usepackage[pagenumbers]{cvpr} 

\usepackage{graphicx}
\usepackage{amsmath}
\usepackage{amssymb}
\usepackage{booktabs}
\usepackage[labelfont=bf]{caption}

\usepackage[pagebackref,breaklinks,colorlinks]{hyperref}

\usepackage[capitalize]{cleveref}
\crefname{section}{Sec.}{Secs.}
\Crefname{section}{Section}{Sections}
\Crefname{table}{Table}{Tables}
\crefname{table}{Tab.}{Tabs.}

\def\cvprPaperID{44} 
\def\confName{CVPR Workshop XAI4CV}
\def\confYear{2022}

\begin{document}

\title{Towards ML Methods for Biodiversity: A Novel Wild Bee Dataset and Evaluations of XAI Methods for ML-Assisted Rare Species Annotations}

\author{\textbf{Teodor Chiaburu}\\
Berlin University of Applied Sciences\\
{\tt\small chiaburu.teodor@bht-berlin.de}
\and
\textbf{Felix Bießmann}\\
Berlin University of Applied Sciences\\
Einstein Center Digital Future\\
{\tt\small felix.biessmann@bht-berlin.de}
\and
\textbf{Frank Haußer}\\
Berlin University of Applied Sciences\\
{\tt\small frank.hausser@bht-berlin.de}
}

\maketitle

\begin{abstract}
    Insects are a crucial part of our ecosystem. Sadly, in the past few decades, their numbers have worryingly decreased. In an attempt to gain a better understanding of this process and monitor the insects populations, Deep Learning may offer viable solutions. However, given the breadth of their taxonomy and the typical hurdles of fine grained analysis, such as high intraclass variability compared to low interclass variability, insect classification remains a challenging task.
   There are few benchmark datasets, which impedes rapid development of better AI models. The annotation of rare species training data, however, requires expert knowledge. Explainable Artificial Intelligence (XAI) could assist biologists in these annotation tasks, but choosing the optimal XAI method is difficult. Our contribution to these research challenges is threefold: 1) a dataset of thoroughly annotated images of wild bees sampled from the iNaturalist database, 2) a ResNet model trained on the wild bee dataset achieving classification scores comparable to similar state-of-the-art models trained on other fine-grained datasets and 3) an investigation of XAI methods to support biologists in annotation tasks. All the code and the link to the dataset can be found in our \href{https://github.com/TeodorChiaburu/beexplainable}{GitHub repository}. 
\end{abstract}


\section{Introduction}
\label{sec:intro}

A study published in 2017 \cite{krefeld}, based on manually collected data, revealed a shocking 75\% decline in insect biomass in German nature protected areas over the past 27 years. Steps have already been taken to determine which areas are primarily affected by insect depopulation and which species face a high risk of extinction \cite{dina, insektensommer}.
This is called \textit{insect monitoring} and, given the current availability of technology, the goal is to construct automatic systems able to recognize insect species without needing to trap and kill any insects. 

Automated insect monitoring requires accurate computer vision (CV) models and development of these models, in turn, requires high quality datasets. Building such datasets is difficult, as there are few domain experts (entomologists) that can annotate images of rare species. In an ideal human-machine collaborative setting, a Machine Learning (ML) algorithm could assist human experts annotating images of rare species by automatically annotating some images and referring others to domain experts. 

When working with such complex structures as an insect taxonomic tree, model explainability becomes a requirement rather than a nice-to-have feature. The entomologists delegated to supervise the monitoring procedure need to understand what exactly led their model to the given output.

\section{Related Work}

A key requirement for better XAI methods are quality metrics, so that model explanations can be evaluated and compared to one another. The XAI community has been working arduously in the past years to develop new XAI methods \cite{Guidotti2018} as well as to formalize the notions of \textit{explanation} and \textit{explainability} \cite{doshi2017towards}. Many methods and software packages have been proposed to test XAI methods automatically \cite{Adebayo2018,kindermans2019reliability,Adebayo2020,innvestigate,CLEVR}. Next to automated evaluations of XAI methods, researchers also investigated XAI methods in human-in-the-loop settings, see e.g. \cite{Hase2020}, but it remains unclear to what extent these two types of evaluations are related \cite{biessmann2021quality}.

Here we compare several XAI methods available in open source libraries \cite{tf-explain, innvestigate}, on the basis of localisation and faithfulness metrics \cite{pointing_game, att_loc, CLEVR, auc, topk_int, PFMCD}.


\section{Dataset}

Our dataset consists of roughly 30k images of wild bees scraped from the \href{https://www.inaturalist.org/}{iNaturalist database}. Only images marked with \textit{Research Grade} were downloaded, meaning that at least two out of three iNaturalist observers agreed upon the insect species. Also, all downloaded photos were licensed under the CC-BY-NC copyright (with a few exceptions, where the owners were explicitly asked for their permission). 

We looked for 25 species, that are met most often in Germany. Out of these 25 classes, four of them are particularly difficult to tell apart from one another based solely on images: \textit{Bombus lucorum}, \textit{Bombus cryptarum}, \textit{Bombus terrestris} and \textit{Bombus magnus}. For the experiments described below, these four classes have been compressed into a single \textit{Bombus lucorum} complex, thus reducing the total number of labels to 22.

\begin{figure}[h]
  \centering
  \begin{subfigure}{0.45\linewidth}
    \includegraphics[width=1.0\textwidth]{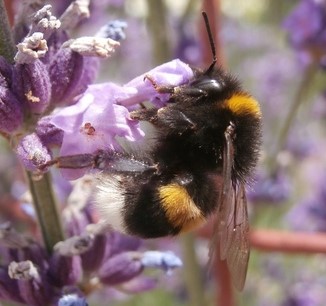}
    \label{fig:lstudio_nomask}
  \end{subfigure}
  \hfill
  \begin{subfigure}{0.45\linewidth}
    \includegraphics[width=1.0\textwidth]{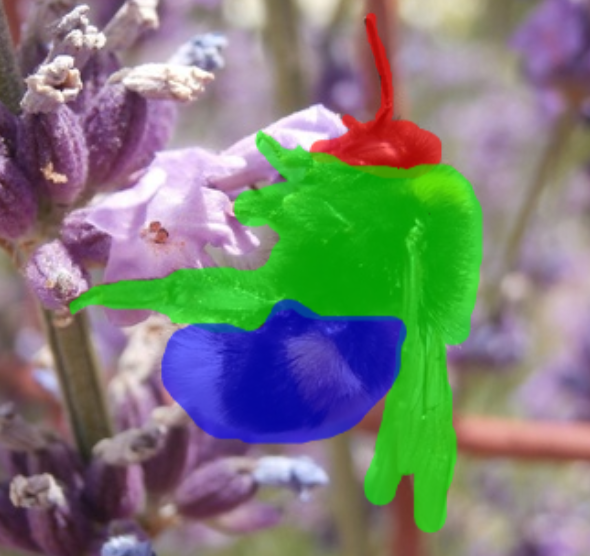}
    \label{fig:lstudio_mask}
  \end{subfigure}
  \caption{Snapshot of a raw (left) and segmented (right) \textit{Bombus lucorum} in Label Studio. Red is for head (including tentacles), green for thorax (including legs and wings) and blue for abdomen.}
  \label{fig:lstudio}
\end{figure}

From the downloaded images, a subset of about 30 images per class (726 total) has been filtered out for further annotation. This subset will be further referred to as the \textit{mini dataset} or \textit{test set}, while the remaining set is the \textit{training-validation set}. Our annotations consist of \textbf{body part segmentation masks} for head, abdomen and thorax. Figure \ref{fig:lstudio} shows an example of a segmented \textit{Bombus lucorum}.


\section{Training and Validation}

\begin{figure*}[h]
    \centering
    \includegraphics[width=1.0\textwidth]{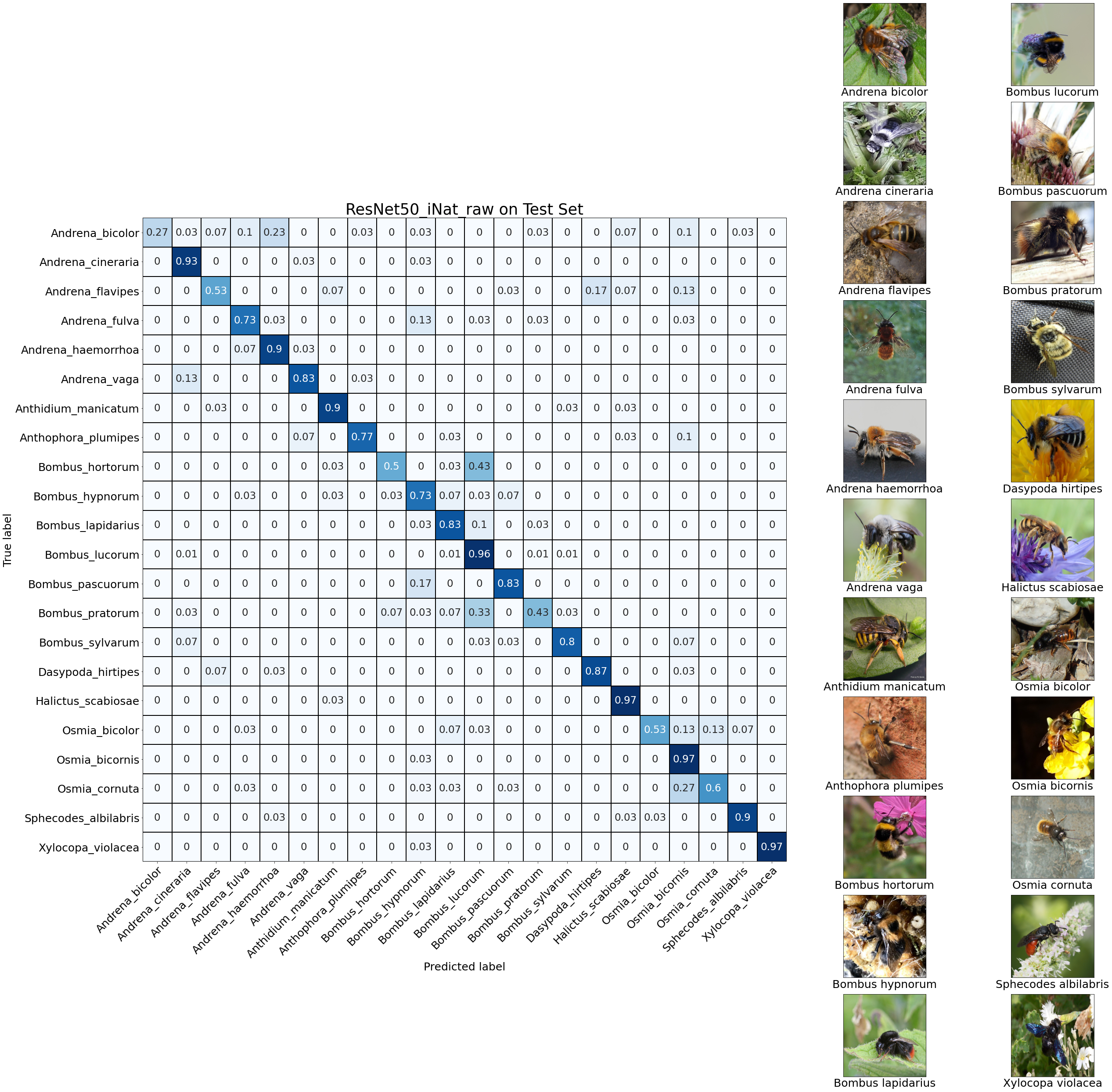}
    \caption{Confusion matrix for the ResNet50 on the test set. The two columns on the right side depict manually chosen prototypical samples for every insect species.}
    \label{fig:conf_mat}
\end{figure*}

Our model is a ResNet50 neural network initialized with the backbone of a similar model \cite{Horn} trained on the iNaturalist Challenge Dataset 2021 \cite{iNat}. A simple linear classifier was added on top. For training, we split the training-validation set into stratified training and validation sets in a 2/3 - 1/3 ratio. We report \textbf{0.78 top-1} and \textbf{0.95 top-3 accuracy} on the test set, which is competitive with similar state-of-the-art fine-grained models \cite{Horn}. Figure \ref{fig:conf_mat} from the Appendix shows the confusion matrix on the test set along with prototypes for every species to ease the understanding of the model's misclassifications.


\section{Experiments}

In order to select an XAI method to support human experts in rare species annotation tasks we compare several established methods with a focus on \textit{post hoc feature attribution based methods}, more specifically \textbf{saliency maps} computed through the \textit{Gradient method}, \textit{Gradient $\times$ Input} \cite{grad} and \textit{Integrated Gradients} \cite{integ}.
In our experiments we use a recently published library \cite{quantus}.
The authors define a simple set of 'desiderata', that is, properties that good XAI methods are expected to have: \textit{faithfulness}, \textit{localisation}, \textit{robustness}, \textit{complexity}, \textit{randomisation} and \textit{axiomatic properties}. For our work, we decided to first investigate the first two and provide initial results. 

\paragraph{Localisation} 

We have used the segmentation masks drawn manually as the ground truth for our model's explanations, measuring to what extent the 'hot pixels' fall into our masks. The results are shown in Table \ref{table:quantus}. We note that our model achieves higher scores on metrics such as Pointing Game and AUC, that evaluate localisation on a coarse level. On the other hand, its performance is rather poor on the other 'finer-grained' metrics, which compute a more in-depth assessment of how well the salient regions overlap with the segmentation masks.

\begin{table}[t]
\centering
\scalebox{0.83}{
\begin{tabular}{ |c||c|c|c|  }
    \hline
     & \multicolumn{3}{|c|}{\textbf{XAI Method}} \\
    \hline
    \textbf{Metric} & Gradient & Grad. x Input & Int. Grad. \\
    \hline
    Pointing Game & \textbf{0.9463} & 0.9421 & 0.9298\\
    Attribution Localisation & 0.2984 & 0.2551 & \textbf{0.3168}\\
    Top-K Intersection & 0.3320 & 0.2807 & \textbf{0.3547}\\
    Relevance Rank Accuracy & 0.3047 & 0.2587 & \textbf{0.3231}\\
    AUC & 0.7157 & 0.6671 & \textbf{0.7297}\\
    \hline
\end{tabular}
}
\caption{Evaluation of three XAI methods from \textit{tf-explain} \cite{tf-explain} according to five localisation metrics from \textit{Quantus} \cite{quantus}. The attribution maps were computed on the whole test set for the ResNet50 w.r.t. the predicted label.}
\label{table:quantus}
\end{table}

\paragraph{Faithfulness}

\begin{figure*}[h]
    \centering
    \includegraphics[width=0.4\textwidth]{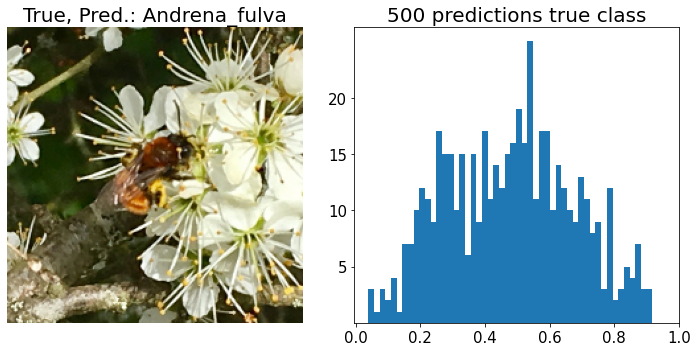}
    \caption{True class probability distribution predicted by the ResNet50 for a sample of the species \textit{Andrena fulva}, after 500 MCD inferences. The probabilities on the y-axis are softmax-normalized. The breadth of the histogram is indicative of the model's predictive uncertainty.}
    \label{fig:mcd_distrib}
\end{figure*}

Inspired by \cite{PFMCD}, we evaluated the faithfulness of the explanations via \textit{Pixel Flipping (PF)} in conjunction with the model's prediction uncertainty via \textit{Monte-Carlo Dropout (MCD)}. By activating the Dropout layer during inference, the network predicts class probability distributions instead of pointwise estimations (Figure \ref{fig:mcd_distrib} from Appendix).
By creating an explanation for every MCD point-prediction of the probability distribution, we get (for every class)  a distribution of pixelwise relevance scores throughout all the generated explanations yielding a diagram of averaged accuracy over percentage of flipped pixels. We would expect good XAI methods to exhibit a rapid and steep decrease in the accuracy curves, as more and more 'relevant' pixels are flipped;
moreover, the saliency flipping curves should fall below the random flipping curve already in the early flipping iterations. Nonetheless, we observe this behaviour only for a couple of classes whereas for most classes, the  saliency flipping does not appear to be more efficient than random flipping (Figures \ref{fig:q_maps} and \ref{fig:pfmcd} in Appendix).


\section{Conclusion and Outlook}

Our results suggest that evaluation of XAI methods remains challenging in niche domains such as rare species classification. The heterogeneity of quality metrics across different XAI methods indicates that choosing the best XAI method for improving annotation experiences for human experts might require further evaluations with human-in-the-loop experiments. We hope that the model, the dataset and the evaluations provided in this work will help researchers to develop models, XAI methods and ML-assisted annotation tools to support entomologists and ultimately improve our understanding of biodiversity.


\section*{Acknowledgements}

We thank Christian Schmid-Egger for providing the list of 25 wild bee species. T.C. and F.H. have been supported within the project KInsecta by German BMUV (67KI2079B).


{\scriptsize
\bibliographystyle{ieee_fullname}
\bibliography{PaperForArxiv}
}


\section*{Appendix}

The following supplementary material includes additional plots and visualizations to support the results of the experiments conducted in this work.

\begin{figure*}[h]
    \centering
    \includegraphics[width=1.0\textwidth]{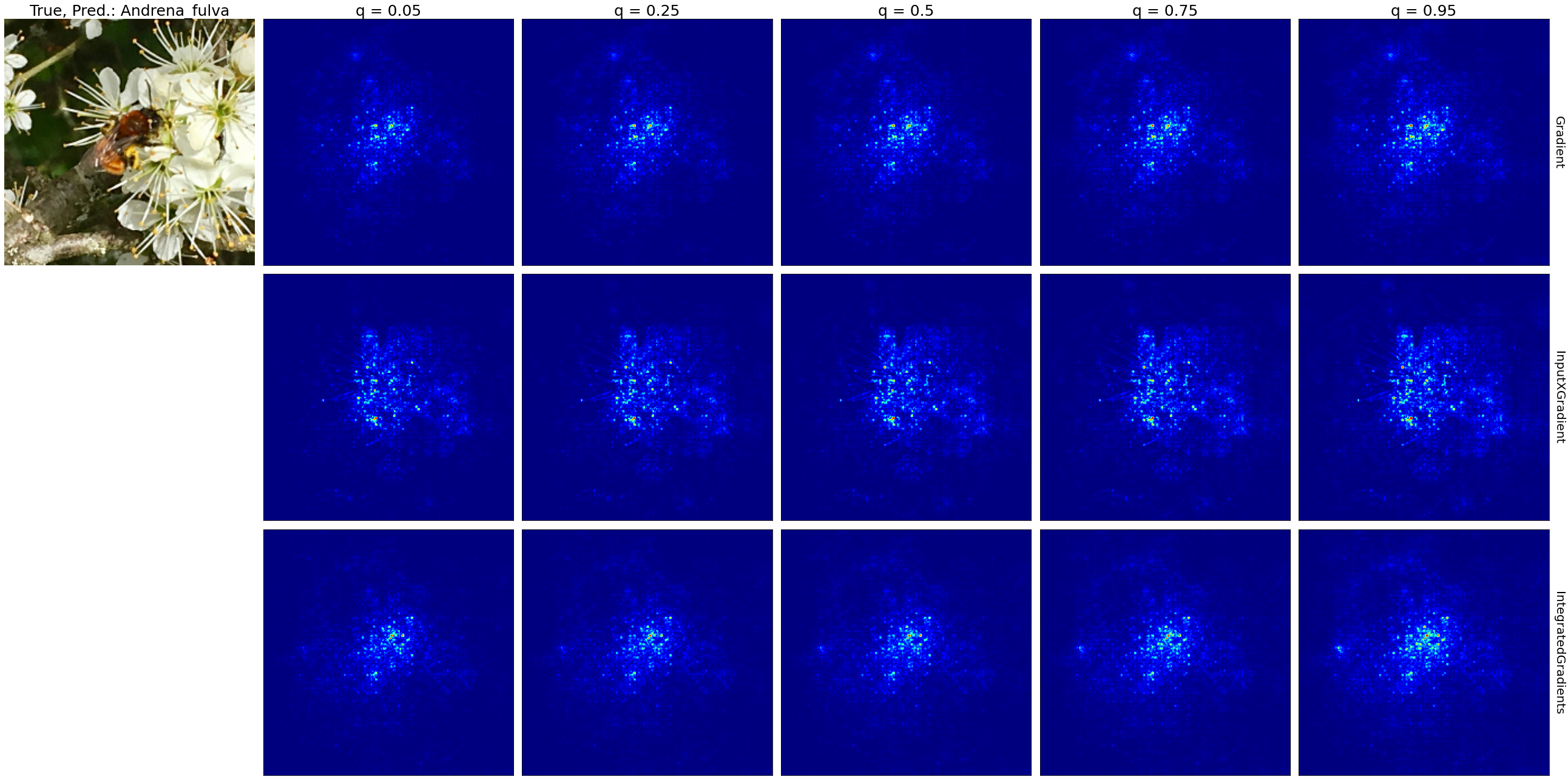}
    \caption{Quantile saliency maps computed for three XAI methods on the basis of the predictions drawn in Figure \ref{fig:mcd_distrib}. For every new (Monte-Carlo) prediction, a new saliency maps is constructed by each of the three methods. The pixels that are marked as relevant throughout all the saliency quantiles are supposed to represent stable relevant features.}
    \label{fig:q_maps}
\end{figure*}

\begin{figure*}[h!]
  \centering
  \begin{subfigure}{0.8\linewidth}
    \includegraphics[width=1.0\textwidth]{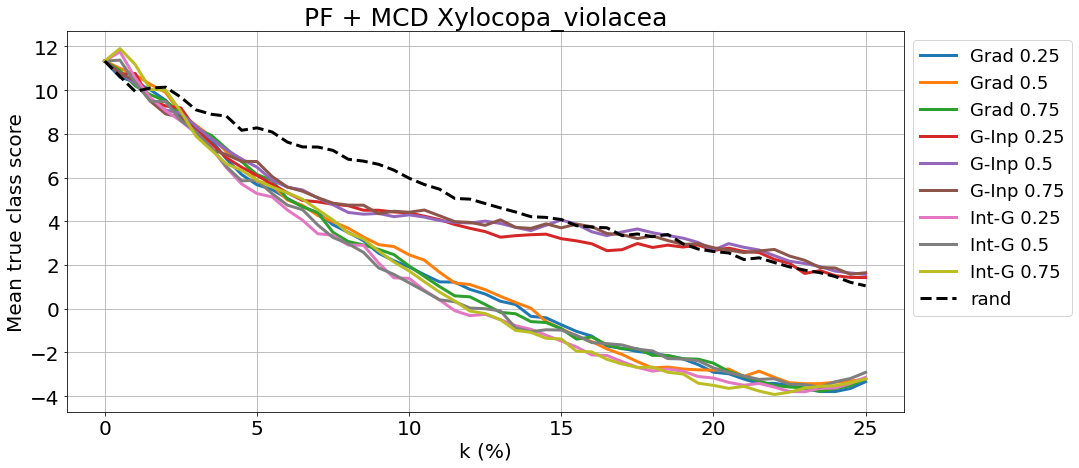}
    \label{fig:pfmcd_good}
  \end{subfigure}
  \hfill
  \begin{subfigure}{0.8\linewidth}
    \includegraphics[width=1.0\textwidth]{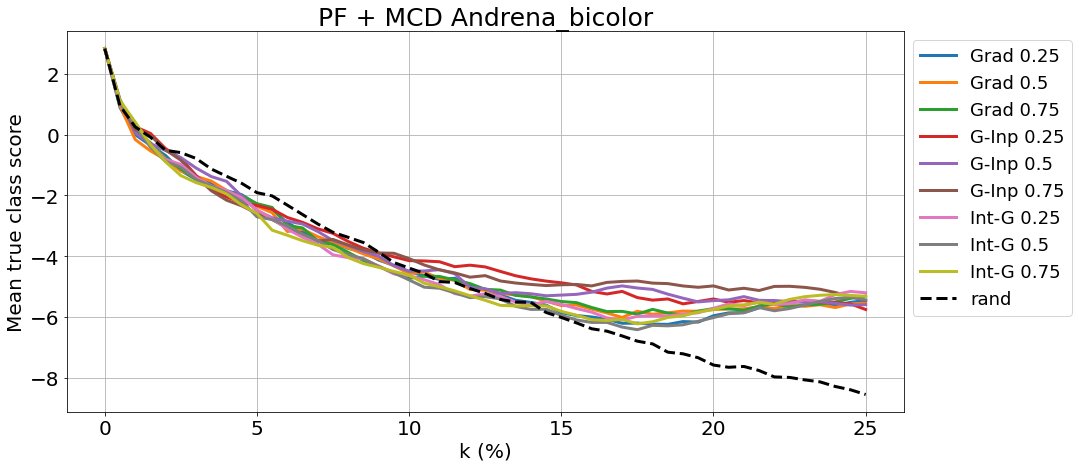}
    \label{fig:pfmcd_bad}
  \end{subfigure}
  \caption{Pixel Flipping for all the samples of the classes \textit{Xylocopa violacea} and \textit{Andrena bicolor} on the test set. The relevant pixels were given by quantile saliency maps for quantiles $q=0.25, \, 0.5, \, 0.75$ generated by three XAI methods (Grad: gradient; G-Inp: gradient $\times$ Input; Int-G: Integrated Gradients) on the ResNet50 with activated MC-Dropout for 100 iterations. The quantile maps (Figure \ref{fig:q_maps}) serve here as markers of relevant features in the image. By iteratively masking (or \textit{flipping}) these relevant pixels, we can judge how important they truly are for identifying the correct class. The greater their relevance, the steeper the decrease in the prediction scores.}
  \label{fig:pfmcd}
\end{figure*}

\end{document}